\documentclass[10pt,twocolumn,letterpaper]{article}

\usepackage{iccv}
\usepackage{times}
\usepackage{epsfig}
\usepackage{graphicx}
\usepackage{amsmath}
\usepackage{amssymb}

\usepackage{algorithm}
\usepackage{algorithmic}
\usepackage[font=small,labelfont=bf]{caption}
\usepackage{subcaption}
\usepackage[dvipsnames]{xcolor}
\usepackage[title,toc]{appendix}
\usepackage{enumitem}
\usepackage{bbm}
\usepackage[numbers,sort&compress]{natbib}

\usepackage{booktabs}
\usepackage{multirow}
\usepackage{array}
\newcolumntype{x}[1]{>{\centering\arraybackslash\hspace{0pt}}p{#1}}
\usepackage{comment}

\usepackage[pagebackref=true,breaklinks=true,letterpaper=true,colorlinks,bookmarks=false]{hyperref}

\iccvfinalcopy 

\def\iccvPaperID{1039} 

\ificcvfinal\fi

\begin{document}

\title{\vspace{-0.3cm}Social NCE: Contrastive Learning of Socially-aware Motion Representations\vspace{-0.2cm}}

\author{
Yuejiang Liu, Qi Yan, Alexandre Alahi\\
École Polytechnique Fédérale de Lausanne (EPFL)\\
{\tt\small\{firstname.lastname\}@epfl.ch}
\vspace{-0.3cm}
}

\maketitle
\ificcvfinal\thispagestyle{empty}\fi

\begin{abstract}

Learning socially-aware motion representations is at the core of recent advances in multi-agent problems, such as human motion forecasting and robot navigation in crowds.
Despite promising progress, existing representations learned with neural networks still struggle to generalize in closed-loop predictions (e.g., output colliding trajectories). 
This issue largely arises from the non-i.i.d. nature of sequential prediction in conjunction with ill-distributed training data.
Intuitively, if the training data only comes from human behaviors in safe spaces, i.e., from ``positive" examples, it is difficult for learning algorithms to capture the notion of ``negative" examples like collisions.
In this work, we aim to address this issue by explicitly modeling negative examples through self-supervision:
(i) we introduce a social contrastive loss that regularizes the extracted motion representation by discerning the ground-truth positive events from synthetic negative ones;
(ii) we construct informative negative samples based on our prior knowledge of rare but dangerous circumstances.
Our method substantially reduces the collision rates of recent trajectory forecasting, behavioral cloning and reinforcement learning algorithms, outperforming state-of-the-art methods on several benchmarks.
Our code is available at \url{https://github.com/vita-epfl/social-nce}.

\vspace{-0.3cm}

\end{abstract}


\newcommand{\debug}[1]{}


\section{Introduction}

Humans have an instinctive ability to anticipate the future motions of other people while navigating in crowded spaces.
This ability allows us to not only keep a comfortable distance from others but also identify potential dangers or discomforts ahead of time. 
However, building predictive models capable of doing so is challenging.
Recent works have proposed a plethora of neural network-based models \cite{deo_convolutional_2018, vemula_social_2017, ivanovic_trajectron_2019,amirian_social_2019,sadeghian_sophie_2019, huang_stgat_2019, kosaraju_social-bigat_2019, sun_recursive_2020, li_evolvegraph_2020} to learn socially-aware motion representations for multi-agent problems such as trajectory forecasting \cite{alahi_social_2016, lee_desire:_2017, gupta_social_2018, salzmann_trajectron_2020} and robot navigation \cite{chen_decentralized_2016, chen_socially_2017, chen_crowd-robot_2019}.
Yet existing methods still output unacceptable solutions (\textit{e.g.}, collisions) from time to time in closed-loop predictions, which raises significant safety concerns for real-world deployment.

\begin{figure}[t]
        \centering
        \includegraphics[width=\columnwidth]{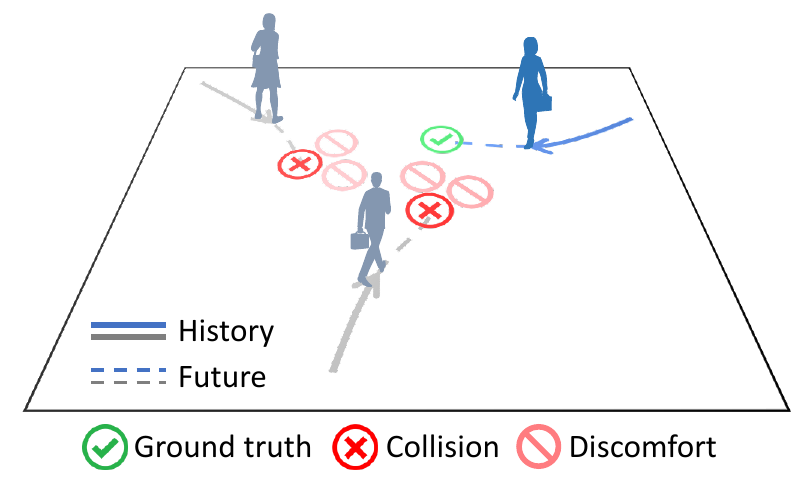}
        \caption{
        Illustration of social contrastive learning. Our method promotes the robustness of neural motion models by means of contrastive representation learning combined with negative data augmentation in the multi-agent context.
        } \label{fig:pull}
        \vspace{-0.5cm}
\end{figure}

One key difficulty in learning robust motion representation stems from the common shortage of critical data.
Very often, the training data is only collected from human behaviors in safe scenarios.
The lack of near-accident examples poses a significant challenge to the discovery of social norms from data.
As a consequence, prediction errors made by the learned model may accumulate over time, gradually create a discrepancy between the training and test state distributions, and eventually cause catastrophic errors \cite{daume_search-based_2009,ross_efficient_2010,codevilla_exploring_2019}.

\begin{figure*}[t]
    \centering
    \begin{subfigure}[b]{0.32\textwidth}
        \centering
        \includegraphics[width=1.0\columnwidth]{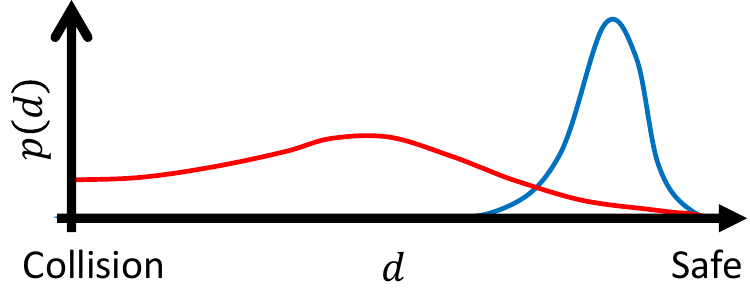}
        \caption{Conventional supervised learning}
    \end{subfigure}
    ~
    \begin{subfigure}[b]{0.32\textwidth}
        \centering
        \includegraphics[width=1.0\columnwidth]{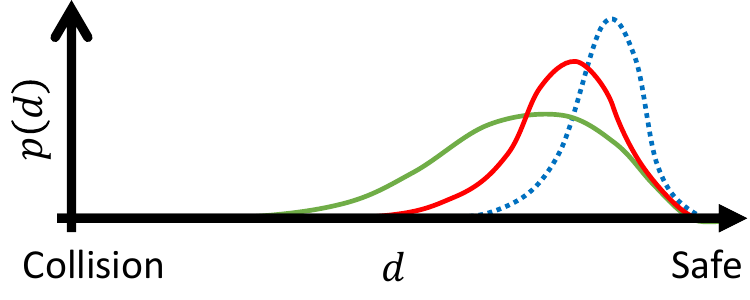}
        \caption{Interactive data collection}
    \end{subfigure}
    ~
    \begin{subfigure}[b]{0.32\textwidth}
        \centering
        \includegraphics[width=1.0\columnwidth]{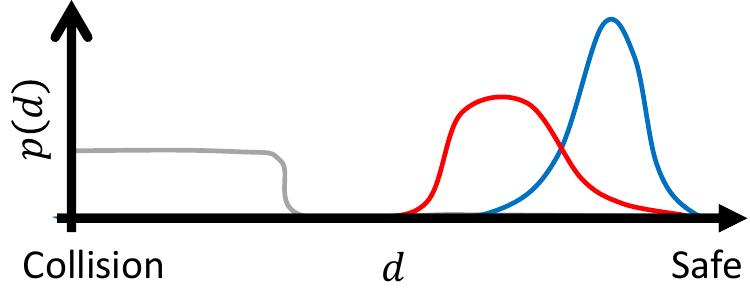} 
        \caption{Our contrastive approach}
    \end{subfigure}
  \caption{Illustration of different learning approaches to socially-aware sequential predictions. (a) The vanilla supervised learning approach often suffers from covariate shift between the training (\textcolor{Blue}{blue}) and test (\textcolor{Red}{red}) data due to the dependence of the state distribution (\textit{e.g.}, separation distance $d$ between agents) on the learned model \cite{daume_search-based_2009, ross_efficient_2010}. (b) Interactive data collection methods \cite{ross_reduction_2011, laskey_dart_2017, ho_generative_2016, de_haan_causal_2019} expand the training distribution to a wider range (\textcolor{Green}{green}) through additional experiments, which are however expensive and even infeasible for forecasting problems. (c) Our social contrastive learning approach augments negative data based on prior knowledge, explicitly informing the learned model about unfavorable states (\textcolor{gray}{gray}) for improved robustness.}
  \label{fig:distribution}
  \vspace{-0.2cm}
\end{figure*}

Previous methods attempt to mitigate this issue through interactive data collections, such as expert queries \cite{ross_reduction_2011, laskey_dart_2017, liu_map-based_2018, sun_deeply_2017, de_haan_causal_2019} and additional experiments \cite{ho_generative_2016, kostrikov_discriminator-actor-critic_2018, wang_random_2019, brantley_disagreement-regularized_2019, reddy_sqil_2019}.
Unfortunately, these methods are not only costly and tedious but often impractical for forecasting problems, since human behaviors can hardly be intervened at scale for data collection.
These shortcomings motivate us to explore an alternative approach: \textit{given a fixed training dataset, can we learn a robust motion representation by exploiting our prior knowledge through self-supervision?}

To this end, we propose a social contrastive learning method built with negative data augmentation (Figure~\ref{fig:pull}).
Our main idea is to promote robust representations by \textit{learning from the opposites} \cite{plato_phaedo_1967}.
Intuitively, an effective way to elucidate the social norms that give rise to the ``positive'' examples is to explicitly portray the opposite ``negative'' examples. We formulate this intuition into an auxiliary contrastive loss, named Social-NCE. It encourages the extracted motion representation to preserve sufficient information for distinguishing a positive event from a set of negative ones.

\vspace{0.02cm}

One crucial component of our method is the design of positive and negative events (states at a specific time step in the future).
Existing contrastive methods in vision and language domains often rely on carefully designed data augmentation for positive pairs while using random samples to form negative pairs \cite{mikolov_distributed_2013, goldberg_word2vec_2014, peters_deep_2018, wu_unsupervised_2018, chen_simple_2020, he_momentum_2020, chen_exploring_2021}. 
Despite its effectiveness for unsupervised pre-training, this common choice is not suitable for motion problems, since it does not bring any extra information than the main task about social norms.
To more explicitly inject our prior knowledge, we introduce a social sampling strategy: construct the positive event from the ground-truth location of the primary agent and the negative events from the regions of other neighbors, given that one location cannot be occupied by multiple agents at the same time.
As illustrated in Figure~\ref{fig:distribution}, our method can be viewed as a form of negative data augmentation \cite{sinha_negative_2020}.
It intentionally informs the model about low-density states through self-supervision, as opposed to laboriously collecting state-action pairs from dangerous scenarios.

\vspace{0.02cm}

We evaluate our method on three tasks: human trajectory forecasting, behavioral cloning, and reinforcement learning for robot navigation in crowds. Experimental results show that our proposed Social-NCE consistently reduces the collision rates and yields new state-of-the-art results on several benchmarks. Our method is model-agnostic and hence can be used as a generic component to promote the robustness of neural motion models.


\section{Related Work}

\begin{figure*}[t]
    \centering
    \vspace{-0.1cm}    
    \includegraphics[width=2.05\columnwidth]{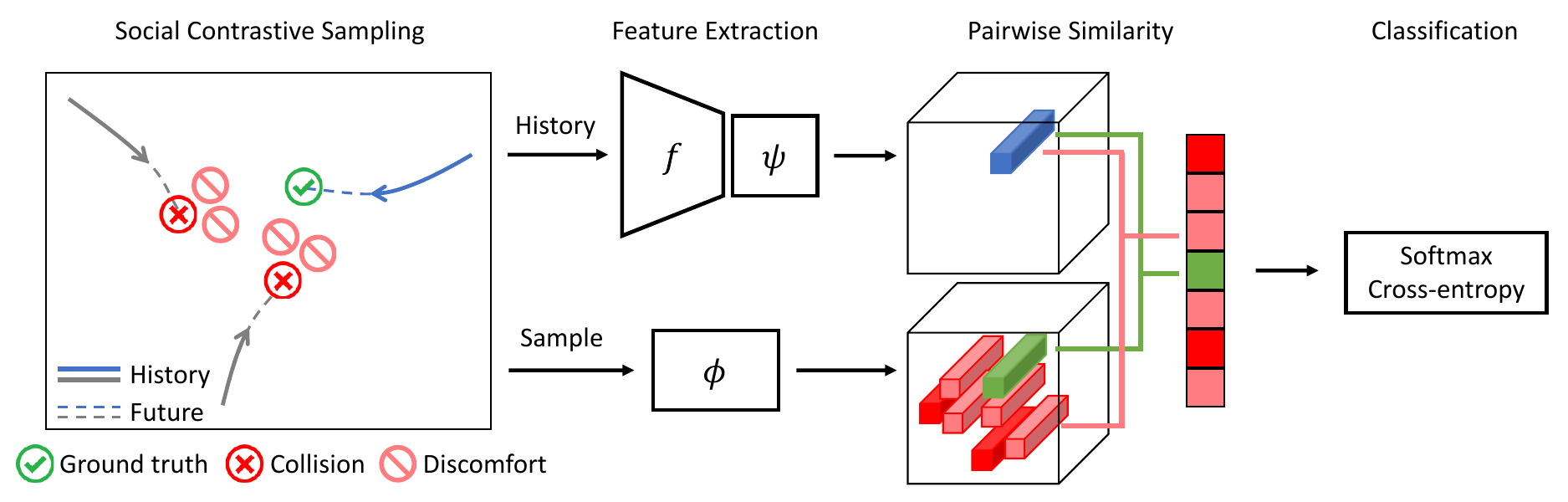} 
    \vspace{-0.2cm}
    \caption{Social contrastive learning in the multi-agent context. Given a scenario that contains a primary agent of interest (\textcolor{Blue}{blue}) and multiple neighboring agents in the vicinity (\textcolor{gray}{gray}), our Social-NCE loss encourages the extracted motion representation, in an embedding space, to be close to a positive future event and apart from some synthetic negative events that could have caused collisions or discomforts.}
    \label{fig:diagram}
    \vspace{-0.2cm}
\end{figure*}

\textbf{Socially-aware Motion Representations.} Human motion in the social context has been traditionally studied based on relative distances and specific rules \cite{helbing_social_1998, mehran_abnormal_2009, van_den_berg_reciprocal_2011, alahi_socially-aware_2014}. While these hand-crafted models have been successfully applied to various tasks \cite{luber_people_2010, zanlungo_social_2011, ferrer_robot_2013,coscia2016point,coscia2018long}, they often struggle to capture the strong interactions among agents in complex scenes \cite{rudenko_human_2020}. Some other methods attempt to learn the patterns of social behaviors from data. Yet early work often suffers from considerable performance drop in densely populated spaces due to limited modeling capacity \cite{trautman_unfreezing_2010, pellegrini_improving_2010}.

More recently, a variety of neural networks have been explored for learning socially-aware motion representations \cite{alahi2017learning,kothari2021its}. Some peculiar neural architecture designs, \textit{e.g.}, feature pooling \cite{alahi_social_2016, gupta_social_2018, deo_convolutional_2018}, attention mechanism \cite{vemula_social_2017, chen_crowd-robot_2019, sadeghian_sophie_2019, huang_stgat_2019}, spatio-temporal graph \cite{ivanovic_trajectron_2019, kosaraju_social-bigat_2019, li_evolvegraph_2020}, latent variable modeling \cite{rhinehart_precog_2019,chai_multipath_2020,choi_drogon_2020}, injecting prior knowledge \cite{bahari2021,Kothari2021cvpr}, have yielded promising results in crowded environments. However, the robustness of these methods remains a central concern. Our work is orthogonal to the design of neural motion models and focused on the learning approach towards robust motion representations.

\textbf{Covariate Shift.} The problem of covariate shift was observed back in \cite{pomerleau_alvinn_1989} and has been a persistent challenge for sequential prediction problems \cite{daume_search-based_2009,ross_efficient_2010}. One practical solution is to actively query experts \cite{ross_reduction_2011,laskey_dart_2017,sun_deeply_2017}, which has been shown effective for behavioral cloning but hardly applicable to forecasting problems \cite{ridel_literature_2018,rudenko_human_2020,kothari2021its}. Inverse reinforcement learning methods \cite{ng_algorithms_2000,abbeel_apprenticeship_2004,ziebart_maximum_2008,ho_generative_2016,kostrikov_discriminator-actor-critic_2018,wang_random_2019,brantley_disagreement-regularized_2019} jointly learn a reward function and the corresponding optimal policy to account for the sequential nature. However, they typically require extensive explorations to solve a reinforcement learning (RL) subproblem \cite{dulac-arnold_challenges_2019}.

Another line of work introduces additional loss terms penalizing the predictions that lead to undesirable events, such as collisions and off-road trajectories \cite{bansal_chauffeurnet_2018, niedoba_improving_2019}. However, these penalties are dependent on the predicted states during training and become utterly ineffective once the model fits the dataset well in late training stages.

Closely related to our work, \cite{luo_learning_2019, zeng_dsdnet_2020} propose to learn a robust value (or cost) function by making use of negative samples. Our method differs from theirs in two key aspects: \cite{luo_learning_2019, zeng_dsdnet_2020} change the task loss that directly affects (and potentially biases) the model output, whereas our goal is to enhance the extracted motion representation without modifications in the main task; they draw negative samples randomly, in contrast, we design a more informed sampling strategy.

\textbf{Contrastive Learning.} Contrastive learning was proposed in \cite{hadsell_dimensionality_2006} to learn an embedding space such that a simple similarity measure can approximate high-level semantic relations. This approach has recently achieved stunning results in a broad spectrum of areas, including computer vision \cite{chen_simple_2020, he_momentum_2020}, natural language understanding \cite{logeswaran_efficient_2018, pagliardini_unsupervised_2018, arora_theoretical_2019}, image synthesis \cite{park_contrastive_2020} and robotics \cite{sermanet_time-contrastive_2018}. Some detailed design choices, such as positive and negative sampling, often play a critical role to the empirical success of contrastive methods \cite{purushwalkam_demystifying_2020, chuang_debiased_2020, kalantidis_hard_2020, robinson_contrastive_2020}. To the best of our knowledge, we are the first to adapt contrastive learning in the multi-agent motion context and explore the sampling method which is unique and critical to socially-aware motion representation learning.


\section{Method}

\label{sec:method}

The robustness of neural motion models has been a long-standing issue \cite{szegedy_intriguing_2013, hendrycks_benchmarking_2018, azadi_discriminator_2018, liu_collaborative_2020}. This issue is particularly concerning in sequential predictions such as behavior forecasting and autonomous navigation.
Very often, the training data only contains ``positive'' examples collected from safe states without any dangerous ``negative'' occurrences.
The severely imbalanced training distribution poses a significant challenge for learning algorithms to truly capture the underlying social norms and generalize to challenging scenarios.

In this section, we present a learning method that aims to tackle this challenge by means of contrastive representation learning combined with negative data augmentation. We will first briefly describe the basic idea of contrastive learning and then present a social contrastive loss. We will finally introduce a sampling strategy tailored for the multi-agent context.

\subsection{Contrastive Representation Learning}

Representation learning typically consists in learning a parametric function (\textit{i.e.}, encoder) that maps the raw data into a feature space to extract abstract and useful information for downstream tasks \cite{bengio_representation_2013}.
Recent contrastive learning methods often adopt the principle of noise contrastive estimation in an embedding space, namely the InfoNCE loss \cite{gutmann_noise-contrastive_2010, dyer_notes_2014, oord_representation_2019}, to train an encoder:
\begin{equation}
    \mathcal{L}_\textbf{NCE} = - \log \frac{\exp( \text{sim}(q, k^+) / \tau)}{ \sum_{n=0}^{N}  \exp( \text{sim}(q, k_n) / \tau)},
    \label{equ:info_nce}
\end{equation}
where the encoded query $q$ is brought close to one positive key $k_0 = k^+$ and pushed apart from $N$ negative keys $\{k_1, \dots, k_N\}$, $\tau$ is a temperature hyperparameter, and $\text{sim}(u,v) = u^\top v/ ({\lVert u \rVert \Vert v \Vert})$ is the cosine similarity between two feature vectors.
It has been shown that minimizing the InfoNCE loss is equivalent to maximize the lower bound on the mutual information between the raw input and the latent representation
\cite{oord_representation_2019}. Moreover, the representations learned by this approach have provable performance guarantees on downstream tasks \cite{arora_theoretical_2019}. The empirical success of this approach often highly relies on the informativeness of positive and negative samples \cite{chen_simple_2020, purushwalkam_demystifying_2020, chuang_debiased_2020, kalantidis_hard_2020, robinson_contrastive_2020, song_multi-label_2020, jabri_space-time_2020}. 

\subsection{Social NCE}

Consider a trajectory forecasting problem in crowded spaces as an example. 
Let $s_t^i = (x_t^i, y_t^i)$ denote the position of agent $i$ at time $t$ and $s_t = \{s_t^1, \cdots, s_t^M\}$ denote the joint state of $M$ agents in the scene. 
Given a sequence of history observations $\{s_1, \cdots, s_t\}$, the task is to predict future trajectories of all agents $\{s_{t+1}, \cdots, s_{T}\}$ until time $T$.
Many recent forecasting models are designed as encoder-decoder neural networks, where the motion encoder $f(\cdot)$ first extracts a compact representation $h_t^i$ with respect to agent $i$ and the decoder $g(\cdot)$ subsequently rolls out its future trajectory $\hat s_{t+1:T}^i$:
\begin{equation}
\begin{split}
h_t^i &= f(s_{1:t},i),  \\
\hat s_{t+1:T}^i &= g(h_t^i).
\end{split}
\end{equation}
To model social interactions among agents, $f(\cdot)$ typically contains two sub-modules: a sequential module $f_S(\cdot)$ that encodes each individual sequence and an interaction module $f_I(\cdot)$ that shares information among agents, \textit{e.g.},
\begin{equation}
    \begin{split}
        z_t^i &= f_S(h_{t-1}^i, s_t^i), \\
        h_t^i &= f_I(z_t, i).
    \end{split}
\end{equation}
where $z_t^i$ is the latent representation of agent $i$ given the observation of its own state at time $t$ and $z_t = \{z_t^1, \cdots, z_t^M\}$.
A variety of architectures have been explored for each modules and validated on accuracy measures \cite{alahi_social_2016, li_end--end_2020, li_evolvegraph_2020}. Nevertheless, their robustness remains an open issue. Several recent works \cite{bansal_chauffeurnet_2018, kothari2021its} have shown that trajectories predicted by existing models often output socially unacceptable solutions (\textit{e.g.}, collisions), indicating a lack of common sense about social norms.

To tackle this challenge, we propose Social-NCE, a variant of InfoNCE adapted to socially-aware motion representations. As illustrated in
Figure~\ref{fig:diagram}, we construct the encoded query and key vectors for the primary agent $i$ at time $t$ as follows:
\begin{itemize}
    \setlength\itemsep{-0.2em}
    \item query: embedding of history observations $q = \psi(h_t^i)$, where $\psi(\cdot)$ is an MLP projection head.
    \item key: embedding of a future event $k = \phi(s_{t+\delta t}^i, \delta t)$, where  $\phi(\cdot)$ is an event encoder modeled by an MLP, $s_{t+\delta t}^i$ is a sampled spatial location and $\delta t > 0$ is the sampling horizon.
\end{itemize}
By tuning $\delta t \in \Lambda$ in a range, \textit{e.g.}, $\Lambda = \{1,\ldots,4\} $,
we can take into account future events in the next few steps simultaneously. Nevertheless, when $\delta t$ is a fixed value, $\phi(\cdot)$ can be simplified as a location encoder, \textit{i.e.}, $\phi(s_{t+\delta t}^i)$. 

In each frame, we draw one positive key and multiple negative keys based on future trajectories in the scene, which we will describe in the next Section~\ref{sec:sampling}. Following~\cite{wu_unsupervised_2018, he_momentum_2020, chen_simple_2020}, we normalize the embedding vectors onto a unit sphere and train the parametric models $f(\cdot), \phi(\cdot), \psi(\cdot)$ jointly with the objective of mapping the positive pair of query and key to similar points, relative to the other negative pairs, in the embedding space:
\begin{equation} \label{eq:snce}
    \mathcal{L}_\textbf{SocialNCE} = - \log \frac{\exp( \psi(h_t^i) \cdot \phi(s_{t+\delta t}^{i,+}, \delta t) / \tau)}{ \sum \limits_{\delta t \in \Lambda}\sum \limits_{n=0}^N \exp ( \psi(h_t^i) \cdot \phi(s_{t+\delta t}^{i,n}, \delta t) / \tau ) }.
\end{equation}
The full training objective is a weighted combination of the conventional task loss, \textit{e.g.}, mean squared error (MSE) or negative log-likelihood (NLL) for trajectory forecasting, and the proposed social contrastive loss:
\begin{equation}
    \mathcal{L}(f, g, \psi, \phi) = \mathcal{L}_\textbf{task} (f, g) + \lambda \mathcal{L}_\textbf{SocialNCE} (f, \psi, \phi),
\end{equation}
where $\lambda$ is a hyper-parameter controlling the emphasis on the Social-NCE loss. 

\begin{figure}[t]
    \begin{subfigure}[b]{0.48\columnwidth}
        \includegraphics[width=1.1\columnwidth]{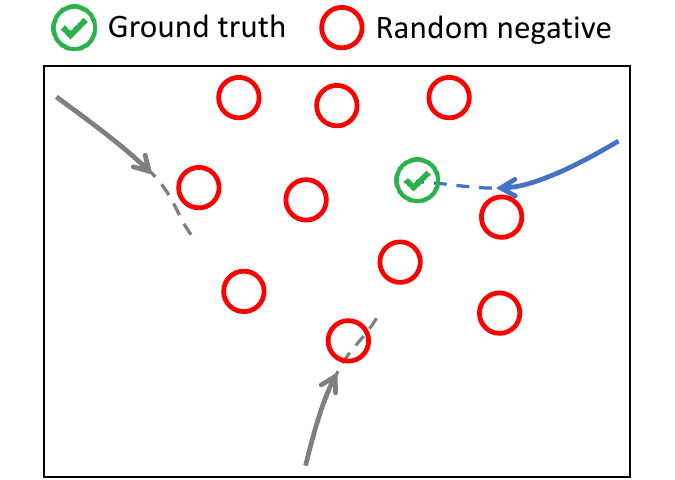}
        \caption{Conventional random negative} \label{fig:sampling_random}
    \end{subfigure}
    \hspace*{-0.1em}
    \begin{subfigure}[b]{0.48\columnwidth}
        \includegraphics[width=1.1\columnwidth]{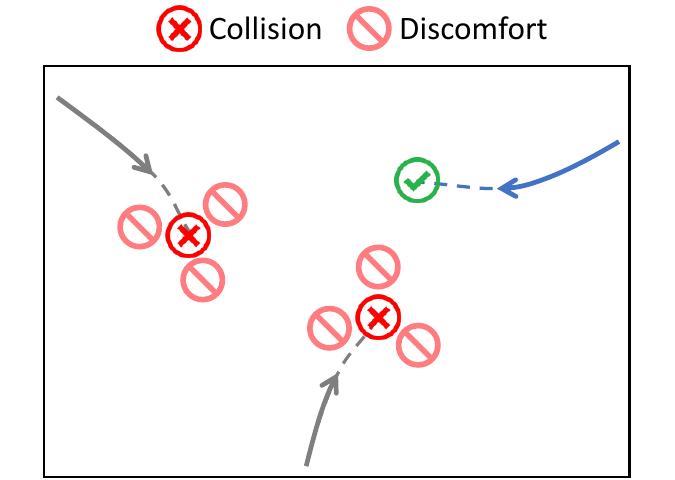} 
        \caption{Our social negative} \label{fig:sampling_social}
    \end{subfigure}\caption{Different negative sampling methods in the multi-agent context. (a) The conventional random sampling method draws negative samples homogeneously scattered in the space, which does not provide much information about social norms. (b) Our social sampling strategy seeks more informative negative samples from the neighborhood of other agents in the future.}
  \label{fig:sampling}
  \vspace{-0.1cm}
\end{figure}

\subsection{Multi-agent Contrastive Sampling} \label{sec:sampling}

One critical design choice of the proposed social contrastive learning lies in the sampling strategy. The recent successes in contrastive learning of visual representations are heavily tied to the use of a large set of negative samples uniformly drawn from the training dataset \cite{mikolov_distributed_2013, goldberg_word2vec_2014, peters_deep_2018, wu_unsupervised_2018, chen_simple_2020, he_momentum_2020}. Unfortunately, this common practice is not suitable for socially-aware motion representation learning. As the main predictive loss already encourages the model to replicate the socially good behaviors from training examples, adding another discrimination task between the correct solution and other randomly scattered negatives cannot provide much extra information about social norms. Worse yet, the random negative sampling, like Figure~\ref{fig:sampling_random}, may contradict to the multimodal nature of the future trajectories and incorrectly penalize plausible solutions.

To effectively incorporate our domain knowledge of socially unfavorable events in the multi-agent context, we propose a social sampling strategy.
As shown in Figure~\ref{fig:sampling_social}, we draw a set of negative samples from the neighborhood of other agents in the future at time $t+\delta t$,
\begin{equation} \label{eq:negative}
    s_{t+\delta t}^{i,n-} = s_{t+\delta t}^{j} + \Delta s_p + \epsilon,
\end{equation}
where $j \in \{1,2,...,M\} \setminus i$ is the index of other agents, and
$\Delta s_p = (\rho \cos{\theta_p}, \rho \sin{\theta_p})$ is a local displacement to account for the social comfort area. For instance, $\rho$ can be the minimum physical distance between two agents and
$\theta_p = 0.25 p \pi, p \in \{0,1,\ldots,7\}$. 
Thus, a total of $N = 8(M-1)$ negative samples are synthesized.
We also add random perturbations to each sampled location $\epsilon \sim \mathcal{N}(0; c_\epsilon \mathbf{I})$, where $c_\epsilon$ is a small constant, \textit{e.g.}, 0.05 [m], to prevent over-fitting.
For positive sampling, we pick a location from the ground truth region of the primary agent $i$ at time $t+\delta t$,
\begin{equation} \label{eq:positive}
    s_{t+\delta t}^{i,+} = s_{t+\delta t}^{i} + \epsilon.
\end{equation}

The key intuition behind our method is that the conventional learning approaches only focus on replicating ``positive'' behaviors in normal states, without the need to understand the consequence of ``negative'' ones. In contrast, by using Social-NCE in tandem with our proposed sampling strategy, we actively enforce the extracted motion representation $h_t^i$ to contain necessary information for identifying the scenarios that could have led to catastrophic outcomes. This subtle but essential difference enables the model to learn a significantly more robust representation from the fixed training dataset.



\section{Experiment}

We empirically validate the proposed Social-NCE on three different tasks: (i) human trajectory forecasting, (ii) imitation learning and (iii) reinforcement learning for robot navigation in multi-agent environments.

On each task, we compare the models obtained by three training methods:
\begin{itemize}
    \setlength\itemsep{-0.2em}
    \item Vanilla: models trained without contrastive loss.
    \item Random: models trained with contrastive loss and the random negative sampling (Figure~\ref{fig:sampling_random}).
    \item Social-NCE (ours): models trained with contrastive loss and the social negative sampling strategy (Figure~\ref{fig:sampling_social}).
\end{itemize}

\begin{table*}[t]
\centering
\small
\begin{tabular}{@{\hspace{0.3cm}}x{1.5cm}@{\hspace{0.8cm}}cc@{\hspace{5mm}}cc@{\hspace{5mm}}c@{\hspace{1.2cm}}cc@{\hspace{5mm}}cc@{\hspace{5mm}}c@{\hspace{0.3cm}}}
\toprule
          & \multicolumn{5}{c}{Social-STGCNN \cite{mohamed_social-stgcnn_2020}}                                  & \multicolumn{5}{c}{Trajectron++ \cite{salzmann_trajectron_2020}}                             \\ \cmidrule(){2-11} 
Dataset         & \multicolumn{2}{c@{\hspace{8mm}}}{Vanilla} & \multicolumn{2}{c@{\hspace{8mm}}}{Ours} & Gain     & \multicolumn{2}{c@{\hspace{8mm}}}{Vanilla} & \multicolumn{2}{c@{\hspace{8mm}}}{Ours} & Gain    \\ \cmidrule(){2-11}  
& FDE {$\downarrow$} & COL {$\downarrow$} & FDE {$\downarrow$} & COL {$\downarrow$} & COL {$\uparrow$} & FDE {$\downarrow$} & COL {$\downarrow$} & FDE {$\downarrow$} & COL {$\downarrow$} & COL {$\uparrow$} \\ \midrule
ETH                   & 1.223                   & 1.33                    & 1.224                   & 0.61                    & \textbf{54.1\%}                   & 0.810                   & 1.16                    & 0.791                   & 0.00                    & \textbf{100.0\%}                 \\
Hotel                 & 0.687                   & 3.82                    & 0.678                   & 3.35                    & \textbf{12.3\%}                    & 0.184                   & 0.84                    & 0.177                   & 0.38                    & \textbf{54.6\%}                   \\
Univ                  & 0.912                   & 9.11                    & 0.879                   & 6.44                    & \textbf{29.3\%}                   & 0.450                   & 3.38                    & 0.435                   & 3.08                    & \textbf{8.9\%}                 \\
Zara1                 & 0.525                   & 2.27                    & 0.515                   & 1.02                    & \textbf{55.1\%}                   & 0.320                   & 0.46                    & 0.330                   & 0.18                    & \textbf{61.5\%}                  \\
Zara2                 & 0.480                   & 6.86                    & 0.482                   & 3.37                    & \textbf{50.9\%}                   & 0.250                   & 1.03                    & 0.255                   & 0.99                    & \textbf{3.3\%}                  \\ \cmidrule(){1-11}
Average               & 0.765                   & 4.70                    & 0.756                   & 2.96                    & \textbf{37.0\%}                   & 0.403                   & 1.37                    & 0.398                   & 0.93                    & \textbf{45.7\%}                  \\ \bottomrule
\end{tabular}
\vspace{-0.1cm}
\caption{Comparison between Social-NCE and the vanilla predictive learning for two recent multi-modal trajectory forecasting models, Social-STGCNN \cite{mohamed_social-stgcnn_2020} and Trajectron++ \cite{salzmann_trajectron_2020}, on the ETH \cite{pellegrini_improving_2010} / UCY \cite{lerner_crowds_2007} datasets. Our method reduces the collision rates of these state-of-the-art models by more than 37\%, while being on par with the vanilla training counterparts in terms of the top-20 final displacement error. Note that we run the official public code to obtain
the baseline results, which are subject to mild differences ($<$ 2\% on average) from the corresponding papers.}
 \label{tab:ethucy}
 \vspace{-0.0cm}
\end{table*}

\begin{table*}[t]
\centering
\small
\begin{tabular}{@{\hspace{0.3cm}}x{1.5cm}@{\hspace{0.8cm}}cc@{\hspace{5mm}}cc@{\hspace{5mm}}c@{\hspace{1.2cm}}cc@{\hspace{5mm}}cc@{\hspace{5mm}}c@{\hspace{0.3cm}}}
\toprule
          & \multicolumn{5}{c}{Social-LSTM \cite{alahi_social_2016}}                                  & \multicolumn{5}{c}{Directional-LSTM \cite{kothari2021its}}                       \\ \cmidrule(){2-11} 
Category  & \multicolumn{2}{c@{\hspace{8mm}}}{Vanilla} & \multicolumn{2}{c@{\hspace{8mm}}}{Ours} & Gain     & \multicolumn{2}{c@{\hspace{8mm}}}{Vanilla} & \multicolumn{2}{c@{\hspace{8mm}}}{Ours} & Gain    \\ \cmidrule(){2-11} 
& FDE {$\downarrow$} & COL {$\downarrow$} & FDE {$\downarrow$} & COL {$\downarrow$} & COL {$\uparrow$} & FDE {$\downarrow$} & COL {$\downarrow$} & FDE {$\downarrow$} & COL {$\downarrow$} & COL {$\uparrow$} \\ \midrule
Avoidance & 1.23         & 13.12         & 1.23       & 10.83       & \textbf{17.5\%} & 1.33         & 12.92        & 1.33       & 10.42       & \textbf{19.4\%} \\
Group     & 0.97         & 7.44          & 0.97       & 4.65        & \textbf{37.5\%} & 1.05         & 6.51         & 1.05       & 5.89        & \textbf{9.5\%}  \\
Overall   & 1.14         & 6.44          & 1.14       & 5.31        & \textbf{17.6\%} & 1.22         & 5.49         & 1.22       & 4.59        & \textbf{16.4\%} \\ \bottomrule
\end{tabular}
\caption{Comparison between Social-NCE and the vanilla predictive learning for two top-ranking uni-modal forecasting models on different interacting categories of the Trajnet++ benchmark \cite{kothari2021its}. Our method reduces the collision rate of the previously most robust model by a clear margin. Note that we obtain all results from the benchmark reports, where FDEs are rounded to two decimal places.}
\vspace{-0.2cm}
\label{tab:trajnet++}
\end{table*}

\subsection{Implementation Details}

In our experiments, we use two different 2-layer MLPs as the projection head $\psi(\cdot)$ and the event encoder $\phi(\cdot)$. We encode the history observations and future events into 8-dimensional embedding vectors.
The distance hyper-parameter $\rho$ is set as 0.2 [m] for trajectory forecasting tasks and 0.6 [m] for robot navigation according to the geometric size of agents in environments. By default, the sampling horizon $\delta t$ is set up to 4 and the temperature $\tau$ is set as 0.1. All models are trained with the Adam optimizer \cite{kingma_adam:_2014}.

\subsection{Trajectory Forecasting} \label{sec:forecasting}

We first evaluate our method on the human trajectory forecasting task. We compare the performances of several forecasting models trained with and without the proposed Social-NCE on the ETH \& UCY benchmark \cite{pellegrini_improving_2010, lerner_crowds_2007} and the Trajnet++ interaction-centric benchmark \cite{kothari2021its}. For a fair and direct comparison with prior work, we implement our learning method in the official public code of each model without any modifications on architectures.

All forecasting models are trained and evaluated in the following setting: given 8 time steps (3.2 seconds) of observations as input, predict future trajectories for 12 time steps (4.8 seconds) for all human agents in the scene. Similar to previous work \cite{gupta_social_2018, zeng_dsdnet_2020, kothari2021its}, we measure the performance of each model with two metrics:
\begin{itemize}
\setlength\itemsep{-0.2em}
    \item Final displacement error (FDE): the Euclidean distance between the predicted output and the ground truth at the last time step.
    \item Collision rate (COL): the percentage of test cases where the predicted trajectories of different agents run into collisions.
\end{itemize}
More experimental details are outlined in the Appendix~\ref{sec:benchmark_forecasting}. 

\subsubsection{ETH \& UCY Benchmark}

We start the evaluation of our method on the ETH and UCY datasets, which contain 5 subsets of real-world pedestrian trajectories widely used in previous work \cite{alahi_social_2016, gupta_social_2018, mohamed_social-stgcnn_2020, salzmann_trajectron_2020}. We consider the following two state-of-the-art multi-modal forecasting models as baselines:
\begin{itemize}
\setlength\itemsep{-0.2em}
    \item Social-STGCNN \cite{mohamed_social-stgcnn_2020}: a spatio-temporal convolutional network with graph representations of pedestrian trajectories.
    \item Trajectron++ \cite{salzmann_trajectron_2020}: a VAE-based recurrent model using element-wise sum to aggregate interactions.
\end{itemize}

Table~\ref{tab:ethucy} shows the experimental results in terms of Top-20 FDE and collision rate. Compared with the conventional predictive learning, our Social-NCE reduces the collision rate by 37.0\% on average for the Social-STGCNN and 45.7\% for the Trajectron++, while remaining competitive with respect to the final displacement error. These results suggest the benefits of our method for boosting the robustness of modern trajectory forecasting models without hurting the accuracy and diversity of endpoint predictions.

\subsubsection{Trajnet++ Benchmark}

We further validate our method on the Trajnet++~\cite{kothari2021its}, an emerging open challenge with an emphasis on interaction modeling. Our evaluation is performed on the following interacting sub-categories:
\begin{itemize}
\setlength\itemsep{-0.2em}
    \item Avoidance: the sub-category where the primary pedestrian avoids others coming from the opposite direction.
    \item Group: the sub-category where the primary pedestrian keeps a close and approximately constant distance with one or more moving neighbors.
    \item Overall: all the scenes with a presence of social interactions, including the ones that are hard to categorize.
\end{itemize}
These interacting categories exclude both \textit{linear} trajectories that often dilute the evaluations of interaction modeling and \textit{non-linear yet non-agent-interacting} trajectories that are strongly affected by the scene context or other unobservable variables, and are therefore particularly suitable for assessing the social awareness of forecasting models.

We use two top-ranking models on the Trajnet++ benchmark as baselines:
\begin{itemize}
\setlength\itemsep{-0.2em}
    \item Social-LSTM \cite{alahi_social_2016}: a LSTM-based model with an interaction module over hidden states of nearby agents.
    \item Directional-LSTM \cite{kothari2021its}: a LSTM-based model with an interaction module sharing velocities of nearby agents.
\end{itemize}
Table~\ref{tab:trajnet++} shows that, similar to the results on the ETH \& UCY benchmark, our social contrastive method yields lower collision rates than the vanilla predictive learning by a clear margin (9.5\%-37.5\%) across all considered sub-categories. In particular, the Directional-LSTM trained with our method clearly outperforms its counterpart, which was previously the most robust model on the public benchmark. It is worth noting that, again, our method does not have a significant impact on prediction accuracy. We conjecture that this is because our method tends to adjust the output trajectories locally instead of changing their global patterns, as reflected by the qualitative examples in the Appendix~\ref{sec:qualitative}.

\subsection{Imitation Learning}
\label{sec:il}

Next, we examine the effectiveness of Social-NCE applied to imitation learning for robot navigation in dense crowds \cite{chen_decentralized_2016, chen_socially_2017, chen_crowd-robot_2019}. We use an open-sourced simulator of crowd navigation \cite{chen_crowd-robot_2019}, where the task for the robot is to navigate through 5 simulated pedestrians and arrive at the target destination with time efficiency. In each time step, the robot receives the observable states of other agents and outputs an action.
We follow the evaluation protocol in \cite{chen_crowd-robot_2019}, which quantifies the performance of a policy using three metrics: navigation time, collision rate, and the accumulated reward as follows:
\begin{equation}
    r(s_t, a_t)= 
\begin{cases}
    -0.25 & \text{if} ~~ d_m^t < 0 \\
    -0.1 + d_m^t/2 & \text{else if} ~~ d_m^t < 0.2 \\
    1 & \text{else if goal is reached} \\
    0 & \text{otherwise} 
\end{cases}
\end{equation}
where $d_m^t$ is the minimum separation distance between the robot and the humans during the time interval $[t-\Delta t,t]$. 

\subsubsection{Behavioral Cloning}

For imitation learning, we collect a demonstration dataset that consists of 5000 simulation episodes using the pre-trained SARL policy in \cite{chen_crowd-robot_2019} as an expert. We train an imitator on the collected data for 200 epochs and evaluate the average performance of the last 10 models saved every 5 training epochs. 

As shown in Table~\ref{tab:bc}, the imitation learning algorithm trained with random negative sampling fails to outperform the vanilla baseline. In fact, it even worsens the learned policy. In contrast, our method (with weight $\lambda = 0.1$) leads to consistently higher reward and lower collision rate. Specifically, our method reduces the collision rate by approximately 69\% compared with the vanilla baseline and attains an average reward of $0.323$, which is highly close to the result of the demonstrator in \cite{chen_crowd-robot_2019}.

\begin{table}[t]
\centering
\small
    \begin{tabular}{ x{1.2cm} | x{1.9cm} @{\hspace{0.5\tabcolsep}} x{1.9cm}@{\hspace{0.5\tabcolsep}} x{2.2cm}}
    \toprule
    Method & Reward {$\uparrow$} & Time {(s) $\downarrow$} & Collision {(\%) $\downarrow$}\\ \midrule
    Vanilla & 0.28 $\pm$ 0.01  & 10.31 $\pm$ 0.07 & 11.11 $\pm$ 1.45 \\
    Random & 0.24 $\pm$ 0.02 & 10.32 $\pm$ 0.12 & 18.60 $\pm$ 4.69 \\
    Ours & \textbf{0.32} $\pm$ 0.01 & 10.33 $\pm$ 0.07 & \textbf{3.40} $\pm$ 1.36 \\
    \bottomrule
    \end{tabular}
\caption{Quantitative results of imitation learning with different methods on a 5k demonstration dataset. 
Higher is better for reward, and lower is better for the other metrics. 
Compared with the vanilla baseline, our method brings down the collision rate by approximately 69\%.} \label{tab:bc}
\vspace{-0.1cm}
\end{table}

\begin{figure}[t]
    \centering
    \begin{subfigure}[b]{0.45\columnwidth}
        \centering
        \includegraphics[width=1.05\columnwidth]{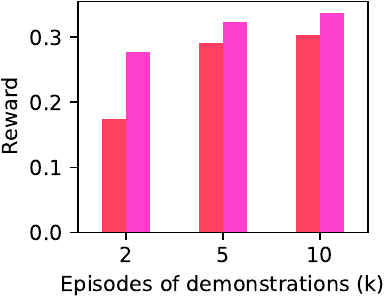}
    \end{subfigure}
    \quad
    \begin{subfigure}[b]{0.45\columnwidth}
        \centering
        \includegraphics[width=1.05\columnwidth]{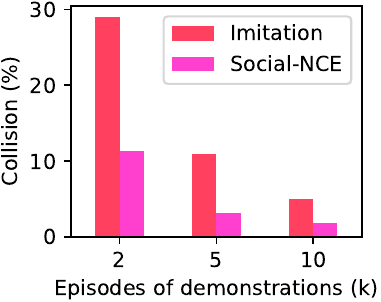}
    \end{subfigure}
  \caption{Social-NCE for imitation learning with different amounts of demonstrations. The conventional behavioral cloning method suffers from a significant performance drop in the low data regime, whereas our method is able to retain much better results thanks to the additional information absorbed from the social contrastive task.}
  \label{fig:efficiency}
  \vspace{-0.4cm}
\end{figure}

\subsubsection{Low-data Regime}

The performance of the standard behavioral cloning approach often degrades substantially when provided with limited demonstrations. We examine the potential of the proposed Social-NCE for data-efficient imitation learning by comparing policies trained on datasets of different sizes. As shown in Figure~\ref{fig:efficiency}, with decreasing amounts of demonstrations, the performance of the vanilla method drops sharply. Notably, the baseline model trained on 2k episodes of demonstrations causes collisions in $29\%$ of test cases. In contrast, our method succeeds in retaining a much higher reward and safety in the low-data regime. For instance, the collision rate of our method with 2k demonstrations is comparable to the baseline with 5k demonstrations. Similarly, our method using 5k training data obtains a higher reward than the counterpart using 10k training data. This result suggests that the learner can absorb a considerable amount of useful information from the designed negative data augmentation, greatly alleviating the information shortage in small training sets.

\begin{figure}[t]
    \centering
    \includegraphics[width=0.7\columnwidth]{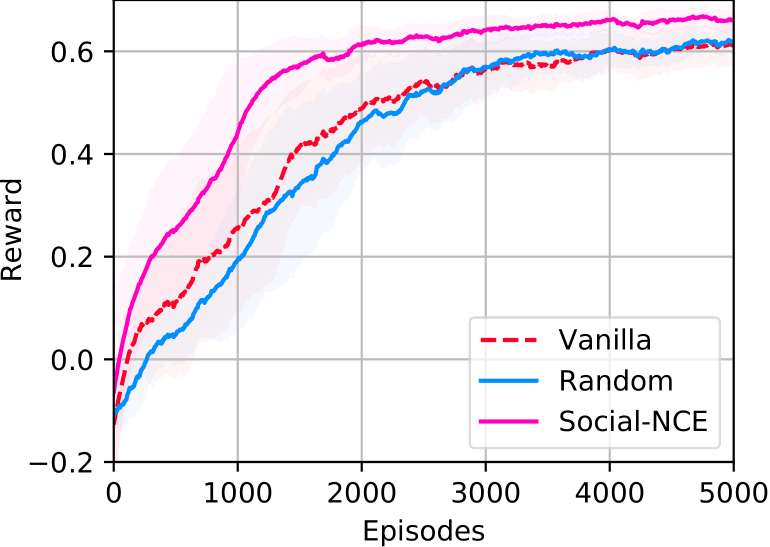}
    \caption{Learning curves of Rainbow DQN \cite{hessel_rainbow:_2017} with different methods for crowd navigation. Results are averaged across 8 random seeds. The shaded area spans one standard deviation. In contrast to the vanilla and random negative counterparts, the agent with Social-NCE is significantly more sample efficient and achieves higher final reward.}
    \label{fig:planning}
    \vspace{-0.1cm}
\end{figure}

\begin{table}[t]
\centering
\small
\begin{tabular}{ x{1.2cm} | x{1.1cm} x{1.1cm} x{1.1cm} x{1.1cm}}
\toprule
\multirow{2}{*}{Method}& \multicolumn{4}{c}{Reward w.r.t. fraction of dataset} \\
& 100\% & 50\% & 25\% & 10\% \\ 
\midrule
Vanilla &   80.1\%\ & 75.2\% & 53.2\% & 14.4\% \\
Random & 81.3\%\ & 71.5\% & 51.7\% & 7.0\% \\
Ours & \textbf{91.6\%} & \textbf{84.6\%} & \textbf{79.2\%}  & \textbf{69.0\%} \\
\bottomrule
\end{tabular}
\caption{Offline RL normalized scores attained by the vanilla rainbow and Social-NCE agents (higher is better).
Normalized score is calculated as: 100 $\times$ (agent score $-$ random play score) $/$ (optimal agent score $-$ random play score).
Our Social-NCE consistently facilitates the recovery of a near-optimal policy and is particularly advantageous in the low-data regime.}
\label{tab:offlineRL}
\vspace{-0.3cm}
\end{table}

\begin{table*}[t]
\vspace{-0.1cm}
\centering
\small
    \begin{tabular}{c | x{2.2cm} | x{2.2cm}@{\hspace{0.6\tabcolsep}} x{2.2cm}@{\hspace{0.6\tabcolsep}} x{2.2cm}@{\hspace{0.6\tabcolsep}} x{2.2cm}@{\hspace{0.6\tabcolsep}} | x{2.2cm} }
    \toprule
    Horizon & Vanilla & 1 & 2 & 3 & 4 & 1-4 \\
    \midrule
    Reward $\uparrow$ & 0.283 $\pm$ 0.008 & 0.281 $\pm$ 0.019 & 0.296 $\pm$ 0.009 & \textbf{0.311} $\pm$ 0.009 & 0.307 $\pm$ 0.012 & \textbf{0.323} $\pm$ 0.005 \\ 
    \midrule
    Time (s) $\downarrow$ & 10.306 $\pm$ 0.068 & 10.345 $\pm$ 0.065 & 10.281 $\pm$ 0.141 & 10.322 $\pm$ 0.134 & 10.348 $\pm$ 0.107 & 10.334 $\pm$ 0.072 \\ 
    \midrule
    Collision (\%) $\downarrow$ & 11.11 $\pm$ 1.45 & 11.24 $\pm$ 3.46 & 9.13 $\pm$ 2.02 & \textbf{5.83} $\pm$ 1.62 & 6.09 $\pm$ 2.26 & \textbf{3.40} $\pm$ 1.36 \\ 
    \bottomrule
    \end{tabular}
\caption{Social-NCE for imitation learning with different sampling horizons. Higher is better for reward, and lower is better for the other metrics. Taking multiple time steps (1-4) into account simultaneously yields better results than a fixed horizon of one time step.} \label{tab:event_ablation}
\vspace{-0.4cm}
\end{table*}

\subsection{Reinforcement Learning}

Finally, we evaluate the proposed Social-NCE for reinforcement learning (RL) algorithms on the crowd navigation task.
We adopt the Rainbow DQN~\cite{hessel_rainbow:_2017}, a state-of-the-art model-free RL method, as baseline and follow the architecture of value-based SARL policy~\cite{chen_crowd-robot_2019} to build the encoder $f(\cdot)$.
To effectively apply Social-NCE to the Rainbow agent, we add a linear layer after the interaction and pooling modules. We also replace the planning module in SARL by dueling and categorical layers, as in the standard Rainbow agent~\cite{hessel_rainbow:_2017}.
In order to isolate the impact of Social-NCE, we use the dense reward function proposed in~\cite{semnani_multi-agent_2020}, which eliminates the necessity of imitation pre-training in \cite{chen_crowd-robot_2019}.

\subsubsection{Off-policy Reinforcement Learning}
We first validate our Social-NCE method in the standard off-policy setting, where an RL agent learns from the replay buffer data gathered over the learning process. We set the temperature as $\tau = 0.2$ and weight as $\lambda=1.0$ for the Social-NCE loss. Figure~\ref{fig:planning} shows the experimental results of each method averaged over 8 random seeds.

The vanilla Rainbow agent reaches a reward value of 0.6 using more than 4000 episodes. In comparison, the agent equipped with our method demonstrates a much higher sample efficiency. It attains the same level of reward in less than 2000 episodes and quickly obtains a collision-free policy thanks to the prior knowledge from the social contrastive task.
Additionally, our method also offers a slight improvement in the final performance.
On the contrary, the random negative sampling is not able to provide any significant performance gain, similar to the experimental results above.

\subsubsection{Offline Reinforcement Learning}

Lastly, we explore the potential of our method in the offline RL setting, in which an agent learns from a static dataset of logged experiences without additional interactions with the environment \cite{levine_offline_2020}. The offline RL setting has attracted rapidly growing attentions due to its tremendous promise for making good use of immerse experience datasets. Nevertheless, most deep reinforcement learning algorithms are highly vulnerable to the distribution mismatch between the policy being trained and the ones used for data collection \cite{lange_batch_2012, fujimoto_off-policy_2019, kumar_stabilizing_2019, islam_off-policy_2019, agarwal_optimistic_2020}.

To verify our method in the offline RL setting, we collect a dataset using the vanilla rainbow agent in the following process: (i) 10k episodes are collected during online RL training from scratch, (ii) multiple free explorations are carried out using online RL checkpoint models trained after $K$ episodes, where $K \in \{500, 1000, 3000, 5000\}$.
Each free exploration contains 5k episodes, and the full dataset is made up of 30k episodes of trials in total. We train an offline Rainbow policy on $N\%$ of the experiences randomly-sampled from the dataset, where $N \in \{10, 25, 50, 100\}$.
We use temperature $\tau=0.2$ and weight $\lambda=0.1$.

Table~\ref{tab:offlineRL} reports the average performance of different offline methods across 10 random seeds. 
Due to the aforementioned distributional shift, no offline methods attain reward scores as high as the online RL algorithm. Nevertheless, our Social-NCE substantially narrows the performance gap and consistently delivers better results than the vanilla rainbow. Notably, the offline policy with our method achieves comparable performance to the best of vanilla baseline using only $25\%$ of the collected data.

\subsection{Ablation: Event Encoder}
To validate the benefits of event encoder $\phi(s_{t+\delta t}^i, \delta t)$, we compare the performance of Social-NCE for imitation learning with different sampling horizons. When the sampling horizon is a fixed value, we use a simplified location encoder $\phi(s_{t+\delta t}^i)$ that only takes the location of a sample as input. Table~\ref{tab:event_ablation} reports the results obtained by using contrastive samples either at a single fixed horizon in a range from 1 to 4 or across all four steps simultaneously. Among the single-step choices, $\delta t = 3$ yields significant performance gains on both reward and collision metrics. On the contrary, $\delta t = 1$ does not provide any improvements in comparison to the baseline due to its short-sightedness. When taking all four steps into account, our method attains the best result, suggesting the importance of establishing social contrastive tasks at multiple horizons.


\section{Conclusion}

In this work, we present a contrastive method for learning socially-aware motion representations.
The proposed Social-NCE loss, combined with our sampling strategy, significantly boosts the performance of recent human trajectory forecasting and crowd navigation algorithms.
Our result suggests that learning from the opposite by means of negative data augmentations can be a promising alternative to the traditional interactive data collection for robust sequential predictions.


\section*{Acknowledgments}

This work is supported by the Swiss National Science Foundation under the Grant 2OOO21-L92326.
We thank Parth Kothari, Yifan Sun, Taylor Mordan, Mohammadhossein Bahari, Lorenzo Bertoni and Sven Kreiss for valuable feedback on early drafts.

{\small
\bibliographystyle{ieee_fullname}
\bibliography{references, extra}
}

\clearpage
\newpage

\appendix

\section{Benchmark Details}

\subsection{Trajectory Forecasting Details} \label{sec:benchmark_forecasting}

We validate our method on two forecasting benchmarks, the ETH \& UCY benchmark and the Trajnet++ benchmark. The former is a \textit{general} benchmark containing pedestrian trajectories in a variety of scenarios. The latter one is based on a curated meta-dataset that consists of \textit{interacting} scenarios selected from several publicly available subsets, such as WildTrack \cite{chavdarova_wildtrack_2018}, L-CAS \cite{sun_3dof_2018} and CFF \cite{alahi_socially-aware_2014}. This benchmark has been used in a series of recent competitions.

Our evaluation protocol follows the previous work \cite{mohamed_social-stgcnn_2020, salzmann_trajectron_2020, kothari2021its}. On the ETH \& UCY datasets, we use the leave-one-out approach, where forecasting models are trained on four sub-datasets and tested on the held-out fifth. On the Trajnet++ dataset, we use the official split of the training and test set. One common feature of recent models like Social-STGCNN and Trajectron++ is that the prediction of the primary agent is only conditioned on the states of neighboring agents up to the observation time $t_o$ but not on any steps from $t_o$ to $t_p$ that have already been predicted, \textit{i.e.}, $s_{t_p+1}^i = f(s_{1:t_p}^i, s_{1:t_o}^{M \backslash i})$. While this design choice accelerates training and inference, it makes the forecasting model unaware of the latest states of the nearby agents and causes notoriously high collision rate at long horizon. As such, our evaluation of collision rate for the Social-STGCNN and Trajectron++ is focused on the first four prediction steps where the models still have access to relatively up-to-date information of the surrounding neighbors. Yet, for the Trajnet++ models that perform fully joint prediction in a recurrent manner, $s_{t_p+1}^i = f(s_{1:t_p}^i, s_{1:t_p}^{M \backslash i})$, we measure the collision rate over the entire prediction horizon.

\subsection{Reinforcement Learning Details} \label{sec:benchmark_rl}

The original SARL policy \cite{chen_crowd-robot_2019} requires a linear motion model as well as imitation pre-training to accomplish the reinforcement learning task from sparse reward feedback. These hand-crafted components, however, introduce extra assumptions over the crowd navigation task and make it hard to analyze the sample efficiency of an RL algorithm. To tease apart the effect of our proposed method, we adopt the following dense reward function \cite{semnani_multi-agent_2020} for the model-free Rainbow algorithm, 

\begin{equation}
\begin{split}
    r(s_t, a_t) &= \alpha(d_g^{t-1} - d_g^{t}) \\ 
    & + \begin{cases}
    -1 & \text{if $d_m^t < 0$} \\
    10d_m^t - 1 & \text{else if $d_m^t < 0.1$} \\
    1 & \text{else if goal is reached} \\
    0 & \text{otherwise}
    \end{cases} 
\end{split}
\end{equation}
where $d_g$ is the Euclidean distance between the robot and its goal, $\alpha = 0.08$ is a control parameter. Other settings are kept the same as  Section~\ref{sec:il}.

\section{Covariate Shift}

To further understand the effect of our learning method on closed-loop sequential predictions, we conduct a detailed analysis of the state distribution at test time on the crowd navigation task with imitation learning. Here, we focus on minimum distance between the robot and the other surrounding agents at each frame and collect a set of robot-human distances from 500 test episodes. 

\begin{figure}[t]
    \centering
    \includegraphics[width=0.7\columnwidth]{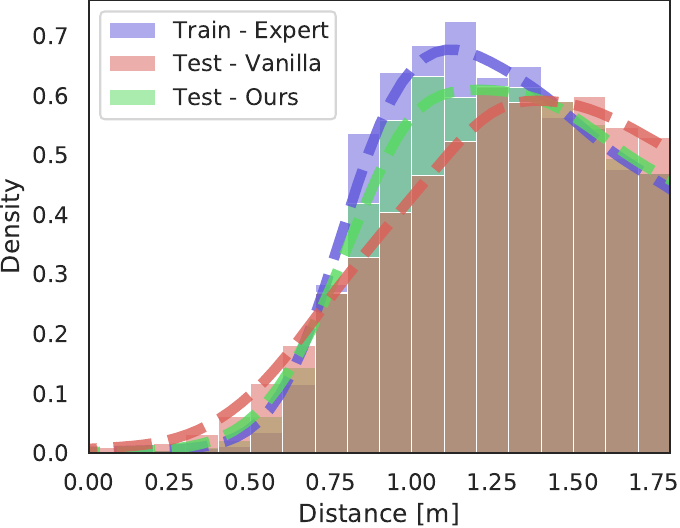}
    \caption{Histogram of the human-robot distance on the crowd navigation task with imitation learning. The vanilla method suffers from the problem of covariate shift in closed-loop sequential predictions, whereas our method results in a much smaller gap between the training and test state distributions.} \label{fig:covariate}
\end{figure}

Figure~\ref{fig:covariate} shows the histogram of human-robot distance with different policies. As expected, the density of short-distance states under the expert demonstrator is close to zero, which reflects the expert's high degree of social awareness as well as confirms the lack of dangerous occurrences in the demonstration data. Compared with the training distribution, the test distribution induced by the model from the vanilla imitation learning yields a clear distinction. It exhibits lower density at the distance around 1.0 [m], but higher density in the dangerous regime, \textit{e.g.}, distance smaller than 0.5 [m]. On the contrary, our method results in a test state distribution almost overlapping with the training one over the dangerous states. These results verify that, given the same distribution of the initial state, the model trained with our method visits dangerous states much less frequently and functions in closed-loop operation much more robustly. 

\section{Qualitative Results} \label{sec:qualitative}

\begin{figure}[t]
    \begin{subfigure}[b]{0.515\columnwidth}
        \centering
        \includegraphics[width=1.0\columnwidth]{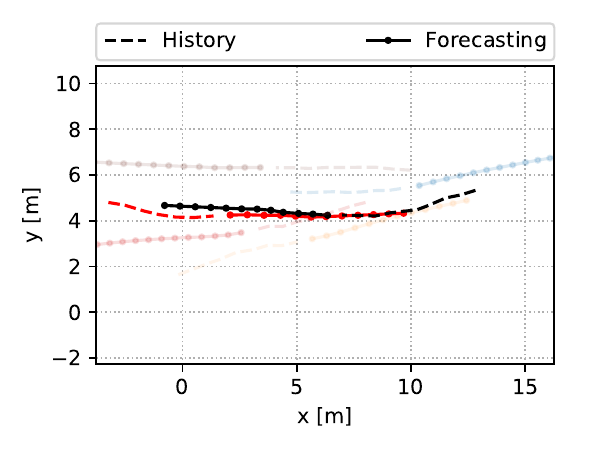}
        \caption{Avoidance - Vanilla}
    \end{subfigure} \hspace*{-1.2em}
    \begin{subfigure}[b]{0.515\columnwidth}
        \centering
        \includegraphics[width=1.0\columnwidth]{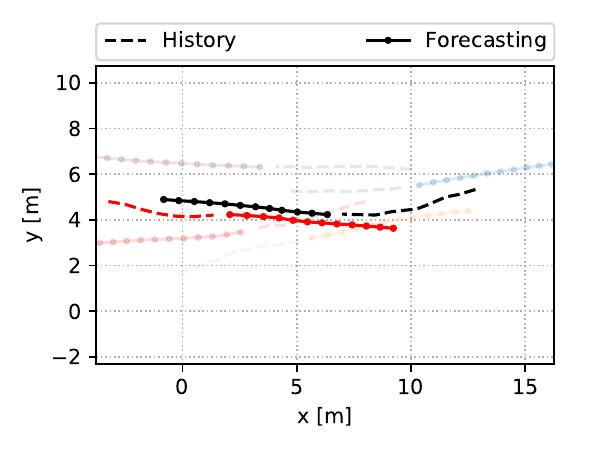}
        \caption{Avoidance - Ours}
    \end{subfigure}
    ~
    \centering
    \begin{subfigure}[b]{0.515\columnwidth}
        \centering
        \includegraphics[width=1.0\columnwidth]{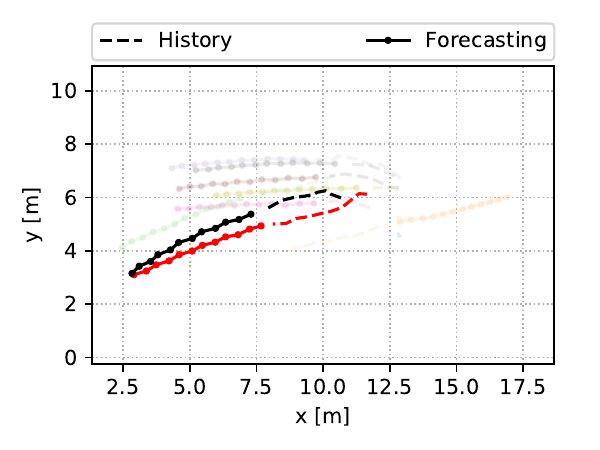}
        \caption{Group - Vanilla}
    \end{subfigure} \hspace*{-1.2em}
    \begin{subfigure}[b]{0.515\columnwidth}
        \centering
        \includegraphics[width=1.0\columnwidth]{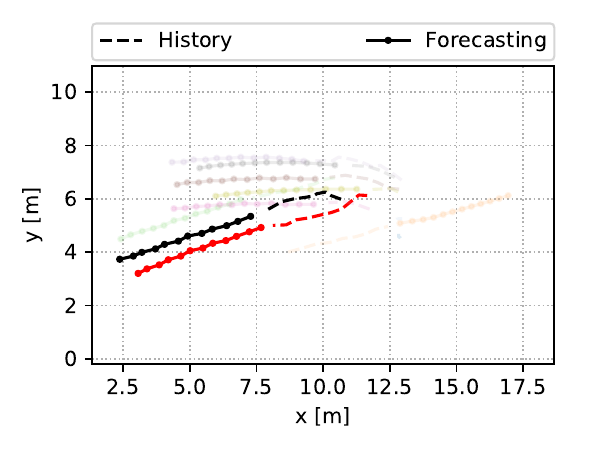}
        \caption{Group - Ours}
    \end{subfigure}
    ~
    \centering
    \begin{subfigure}[b]{0.515\columnwidth}
        \centering
        \includegraphics[width=1.0\columnwidth]{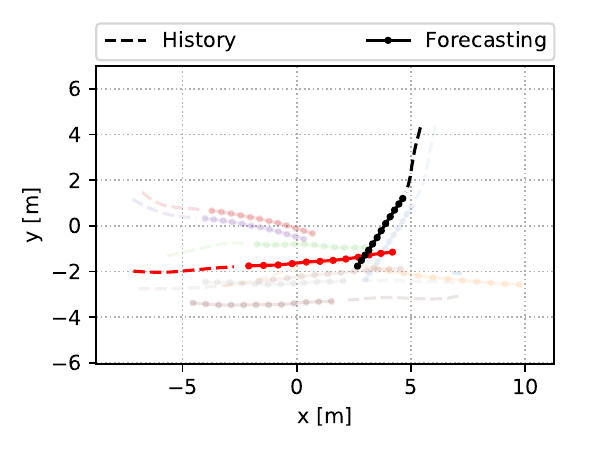}
        \caption{Other - Vanilla}
    \end{subfigure} \hspace*{-1.2em}
    \begin{subfigure}[b]{0.515\columnwidth}
        \centering
        \includegraphics[width=1.0\columnwidth]{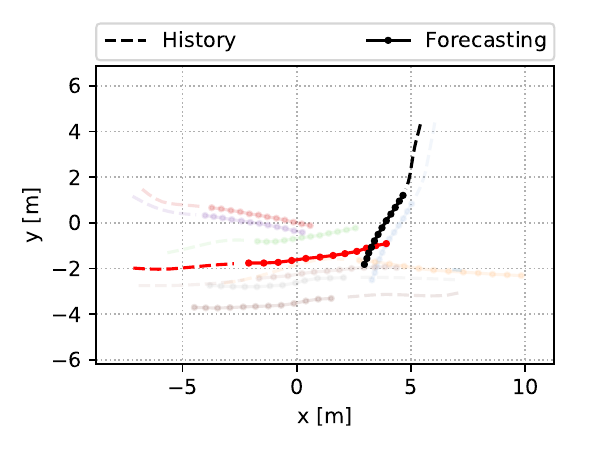}
        \caption{Other - Ours}
    \end{subfigure}
  \caption{Qualitative results of Directional-LSTM \cite{kothari2021its} models trained with different methods in three \textit{interacting} test cases on the Trajnet++ benchmark \cite{kothari2021its}. The vanilla method leads to collisions between the primary (black) and the nearby agent ({\color{red}red}) at the 4th, 12th, and 10th predicted step on the \textit{Avoidance}, \textit{Group} and \textit{Other} case respectively, whereas our method outputs collision-free trajectories over the whole prediction horizon.}
  \label{fig:qualitative}
\end{figure}

In addition to quantitative comparison, Figure~\ref{fig:qualitative} shows the qualitative results of different learning methods in three different interacting scenarios on the Trajnet++ benchmark. The Directional-LSTM \cite{kothari2021its} trained with the vanilla predictive learning outputs colliding trajectories between the primary agent and its neighbors in these dense scenes. In contrast, our method outputs more socially compliant solutions: in the \textit{Group} scenario, our predicted trajectory for the primary agent stays in the middle of two other neighbors at all time steps instead of sliding towards either of them. In the \textit{Avoidance} scenario, our method adjusts the trajectories of both the primary and the opposite agent cooperatively. Similarly, in the \textit{Other} scenario where pedestrians come from almost orthogonal directions, our method jointly twists the trajectories of these interactive agents, enabling each of them to pass the crowded spot smoothly.

\end{document}